\documentclass{article}

\usepackage{arxiv}

\usepackage[utf8]{inputenc} 
\usepackage[T1]{fontenc}    
\usepackage{hyperref}       
\usepackage{url}            
\usepackage{booktabs}       
\usepackage{amsfonts}       
\usepackage{nicefrac}       
\usepackage{microtype}      
\usepackage{lipsum}
\usepackage{graphicx}
\graphicspath{ {./images/} }

\title{Sampling Through the Lens of Sequential Decision Making}

\author{
 Jason Xiaotian Dou \\
  University of Pittsburgh\\
  doucmu@gmail.com
      \And
 Alvin Qingkai Pan\\
  Columbia University\\
  qp2134@columbia.edu
  \And
   Runxue Bao \\
  University of Pittsburgh\\
  runxue.bao@pitt.edu
  \And
  Haiyi Mao \\
  University of Pittsburgh\\
  ham112@pitt.edu
    \And
 Lei Luo\\
  University of Pittsburgh\\
  luoleipitt@gmail.com
  \And
   Zhi-Hong Mao\\
  University of Pittsburgh\\
  zhm4@pitt.edu
}

\begin{document}
\maketitle
\begin{abstract}
Sampling is ubiquitous in machine learning methodologies. Due to the growth of large datasets and model complexity, we want to learn and adapt the sampling process while training a representation. Towards achieving this grand goal, a variety of sampling techniques have been proposed. However, most of them either use a fixed sampling scheme or adjust the sampling scheme based on simple heuristics. They cannot choose the best sample for model training in different stages. Inspired by ``Think, Fast and Slow'' (System 1 and System 2) in cognitive science, we propose a reward-guided sampling strategy called Adaptive Sample with Reward (ASR) to tackle this challenge. To the best of our knowledge, this is the first work utilizing reinforcement learning (RL) to address the sampling problem in representation learning. Our approach optimally adjusts the sampling process to achieve optimal performance. We explore geographical relationships among samples by distance-based sampling to maximize overall cumulative reward. We apply ASR to the long-standing sampling problems in similarity-based loss functions. Empirical results in information retrieval and clustering demonstrate ASR’s superb performance across different datasets. We also discuss an engrossing phenomenon which we name as ``ASR gravity well'' in experiments.
\end{abstract}
\section{Introduction}
Various sampling techniques have been developed to cope with learning problems' high complexity and heterogeneity. In this paper, we approach the sampling problem in a cognitive science-informed way. In cognitive science, Daniel Kahneman develops a theory of ``Thinking, Fast and Slow'' to explain human decision making \cite{kahneman2011thinking,booch2020thinking, saisubramanian2021multi, dou2017impartial,wu2015impartial,douclinical}. In Kahneman's theory, human decisions are supported by two main kinds of capabilities. System 1 is guided by intuition rather than deliberation and gives fast answers to simple questions. On the contrary, system 2 solves more complex questions with additional computational resources, full attention, and sophisticated logical reasoning. Figure 1 gives an overview of the two systems of thinking. A closer comprehension of how humans have evolved to obtain can inspire innovative ways to imbue AI with competencies \cite{booch2020thinking}. \cite{kahneman2011thinking} provides a unified theory aiming to identify the roots of desired human capabilities. 

Current adaptive sampling methods \cite{daneshmand2016starting,mokhtari2016adaptive,mokhtari2017first,lei2016less, mokhtari2019efficient} are similar to System 1 and reinforcement learning \cite{agarwal2019reinforcement, sutton2018reinforcement, kumar2021dr3} is one step closer to System 2. One representative System 1 sampling method is semihard sampling (which yields samples that is fairly hard but not too hard) \cite{wu2017sampling}. While as the sample size grows, relying on relatively simple heuristic based methods to drive learning becomes challenging. System 2 becomes the better approach. We propose System 2 as a reinforcement learning-based sampling approach where a reward function guides the sampling process called Adaptive Sampling with Reward (ASR). We apply the ASR framework to the representation learning sampling problem and achieve better performance than competitive, commonly adopted sampling methods (System 1). We observe an engrossing phenomenon called ``ASR gravity well'' in the reinforcement learning policy initialization step, which concurs the fascinating ``softmax gravity well'' summarized in \cite{mei2020escaping}.
\begin{figure*}[t]
    \centering
    \includegraphics[scale=0.5]{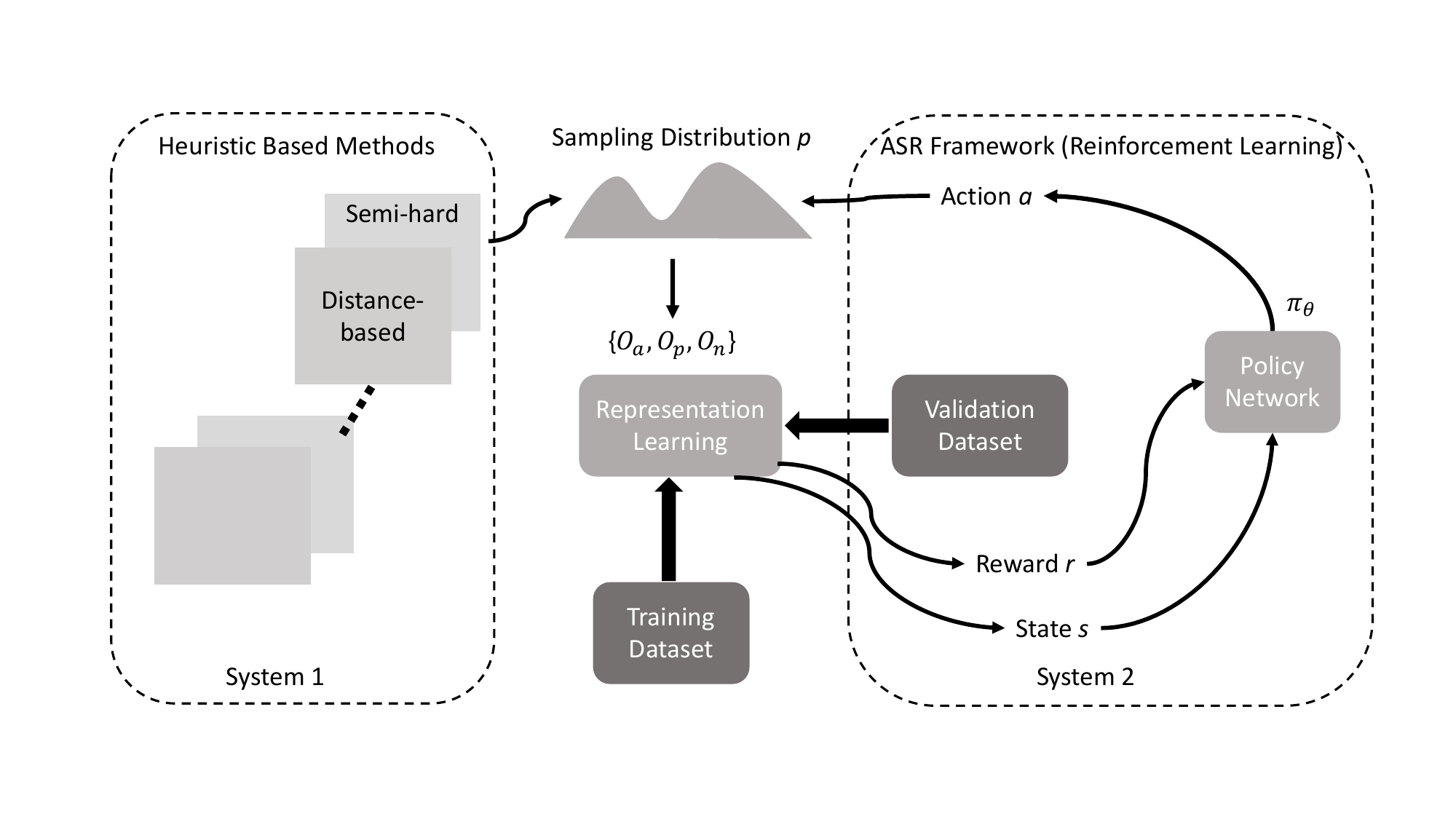}
    \vspace{-1cm}
    \caption{Overview of System 1 vs System 2 approaches to sampling: heuristics based methods like semi-hard sampling and distance based sampling represent System 1, the ASR approach (Reinforcement learning) we propose represents System 2. A comprehensive evaluation of different heuristic based methods can be found in \cite{wu2017sampling}. In Section 4 we give detailed explanation of the ASR framework.}
    \label{fig:arch}
\end{figure*}

We organize the rest of the paper as the following: in Section 2, we review some of the classic and recent work on the applications of reinforcement learning, the interplay between reinforcement learning and representation learning, and adaptive sampling methods. Then in Section 3, we detail reinforcement learning's basic formulation and the representation learning task metrics NMI, Recall, and F1, which we use to construct the reward structure in Section 4. Later in Section 4, we elaborate the System 2 we design. We first explain how we model the sampling process in representation learning as a reinforcement learning problem. Then we present the policy gradient theorem and our way to optimize the policy objective \cite{jin2021pessimism}. Section 5 compares the two systems: the ASR approach with the widely used semi-hard and distance-based sampling strategies on two representation learning tasks, information retrieval, and clustering. We demonstrate the ASR framework's superiority empirically over three benchmark datasets. In Section 5, we provide a social impact statement and discussion, and finally, in Section 6, we conclude our research and suggest several intriguing questions for future research endeavors.
\section{Related Work}
\subsection{Reinforcement Learning} There has been a flurry of recent work on reinforcement learning methodologies and applications \cite{singh2021parrot,shen2019hessian,mao2020near, zhang2019global,wang2019neural,liu2021regularization,lee2020weaklysupervised,yan2020does, wang2021desiderata}. One of the most phenomenal applications of reinforcement learning, which has made a tremendous impact on the AI community and beyond, is the application of Go \cite{silver2016mastering}. Furthermore, \cite{laskin2021cic} works on RL for unsupervised skill discovery. \cite{nasiriany2021augmenting} applies reinforcement learning with behavior primitives for diverse manipulation tasks. \cite{gao2020development} formulates the protein folding as a reinforcement learning problem. \cite{huang2021adarl} proposes a framework adaptive reinforcement learning, called AdaRL, that adapts reliably and efficiently to changes across domains with a few samples from the target domain, even in partially observable environments.

Above scenarios often consider single agent setting. Recently mean field theory \cite{flyvbjerg1993mean} has been used to aggregate agents to address the multi-agent reinforcement learning problems \cite{subramanian2021partially,subramanian2021decentralized, subramanian2022multi}. Learning in Multi-agent reinforcement learning is fundamentally difficult since agents not only interact with the environment but also with each other. Traditional Q-learning considers other agents as the environment fails in multi-agent setting and makes the learning unstable: one agent's policy change will affect other's. Specifically, mean field reinforcement learning addresses the above problem by learning in a stochastic games with a large number of agents, where the empirical mean action is used as mean field. And each agent uses the mean field to update Q-function and policy \cite{yang2020mean,subramanian2021decentralized}.
\subsection{Interplay between Reinforcement Learning and Representation Learning} 
Representation learning allows an AI system to automatically discover the representations needed for feature detection or classification from raw data. Due to its effectiveness in real-world tasks, intensive efforts have been put to design various representation learning algorithms in recent years. Questions in representation learning include appropriating objectives for learning good representations, for computing
representations (i.e., inference), and the geometrical connections between representation learning \cite{mao2022coem, bengio2014representation,maua2012anytime,dou2022optimal,https://doi.org/10.48550/arxiv.1710.00273, wu2022fast,10.1145/3568113.3568124}. \cite{wang2015deep} considers learning representations (features) in the setting in which we have access to multiple unlabeled views of the data for learning while only one view is available for downstream tasks. While \cite{hassani2020contrastive}  introduces a self-supervised approach for learning node and graph level representations by contrasting structural views of graphs. \cite{donahue2019large} shows that progress in image generation quality translates to substantially improved representation learning performance and proposes a large scale adversarial representation learning approach. In addition, representation learning is prominent in machine learning applications in computational biology \cite{paudel2022serologic, flores2021deep, meng2021mimicif,lopez2018deep}, natural language processing \cite{ijcai2021-574, sun2020improving,majewska2021bioverbnet} and computer vision \cite{schroff2015facenet} and computational social science \cite{10.1145/3447548.3467391}. 

Using reinforcement learning in high dimensional data can be arduous and tedious, where the training may not make significant progress without excessive hyperparameter tuning. Indeed, conventional reinforcement learning methods are demonstrated to be restricted by computational and data efficiency issues without a well-learned representation \cite{goel2018unsupervised, shelhamer2016loss, chen2021decision}. Therefore, it's not surprising that numerous work has been done in the past that utilize latent space representations, such as
\cite{dou2022optimal,lee2019stochastic, zhan2020framework, nair2018visual, ghosh2018learning, gelada2019deepmdp}.

In particular, stochastic Latent Actor-Critic (SLAC) \cite{lee2019stochastic} is an enhanced method on top of the classic Soft Actor-Critic (SAC) \cite{haarnoja2018soft} that extends to partially observed Markov Decision Process (POMDP) \cite{pmlr-v115-salmon20a,poupart2013vectorspace}. The intuition is to optimize the joint likelihood of observation (state) space and the optimality of each time step conditioned on past actions given by maximum entropy policies. The exact objective is to maximize the evidence lower bound (ELBO) of the logarithm of the joint likelihood computed from the posterior of the factorization of the variational distribution. By explicitly learning latent representations, reinforcement learning from high-dimensional objects is accelerated, especially for long-horizon continuous control.
\subsection{Adaptive Sampling}
The adaptive sample scheme exploits the specific properties of empirical risk minimization problems to improve the convergence guarantees for
traditional optimization methods. \cite{daneshmand2016starting} proposes two types of sample-size schemes: linear and alternating. Linear means we start from sample size $n$ and perform $2n$ steps. Alternating means every other iteration we sample a new data point, which is added to the set, and there is also update on the fresh sample. Afterwards series of work have extend the framework to different optimization methods and problem settings \cite{daneshmand2016starting,mokhtari2016adaptive,mokhtari2017first,lei2016less, mokhtari2019efficient}.
\section{Problem Formulation and Preliminaries}
\subsection{Reinforcement Learning Formulation}
In reinforcement learning, a learning agent is embedded in an environment in such a way that it can discriminate state space $S$ and implement actions in the action space $\mathcal{A}$\cite{kaelbling1993learning}.
$\forall s \in S$, the agent can choose an available action $a \in \mathcal{A}$ and gains a immediate reward $r(s,a)$ which follows a distribution $R$ given $s\in S$ and $a \in \mathcal{A}$. The environment inherits transition probabilities $P(s', r | s, a)$ that determines the probability of the agent going to $s'$ and gains $r$ by taking $a$ from $s$. 
Now the objective is to learn an optimal policy where the remaining decisions must constitute an optimal policy with regard to the state resulting from the first decisions regardless of the initial state and decisions \cite{bellman1954theory}.
Hence, the agent takes a trajectory with joint probability distribution given by the optimal policy $\pi^*$ so that it maximises the expected accumulative reward $\mathbb{E}[\sum_{t=0}^{\infty} \eta^t r_t]$, where the agent starts at $s_0$ and $\eta \in (0,1]$ is the discount factor \cite{agarwal2019reinforcement}.

Reinforcement learning and contextual bandits are two widely studied algorithms for sequential decision making problems. Reinforcement learning is usually perceived to be more difficult than contextual bandits due to long horizon and state-dependent transitions. Interestingly, \cite{zhang2021reinforcement} shows that reinforcement learning poses little additional difficulty on sample complexity compared to contextual bandit.
\subsection{Evaluation Metrics as Reward} 
In representation learning,  Recall@n(1,2,4), F1, and Normalized Mutual Information \cite{schroff2015facenet, wu2017sampling} are benchmarks for clustering and retrieval tasks. We use them as basis of reward function in reinforcement learning. Thus, we need to give a brief introduction to these metrics.
\begin{itemize}
    \item \textbf{Normalized Mutual Information (NMI)} We use NMI to measure the clustering quality. Normalization of the Mutual Information (MI) score to scale the results between 0 (no mutual information) and 1 (perfect correlation).
    \item \textbf{Recall at k} We calculate Recall@k for retrieval tasks. Specifically, for each query, top-k nearest objects are returned, then the recall score will be calculated. The same class label as the query positive and the others negative.
    \item \textbf{F1 score} The F1-score measures the harmonic mean between precision and recall which is a commonly used retrieval metric, placing equal importance on both precision and recall.
\end{itemize}

\section{Algorithmic Framework: Adaptive Sampling with Reward (ASR)}
\subsection{Representation Learning and Similarity based Loss Functions}
In representation learning, we use $\phi_{i}:=\phi\left(O_{i}; \zeta\right)$ to denote a D-dimensional embedding of an object $O_{i}$ with $\phi\left(O_{i} ; \zeta\right)$ being represented by a deep neural network parameterized by $\zeta$ \cite{schroff2015facenet,lee2019stochastic,wu2017sampling, roth2020revisiting}. This representation makes using similarity based loss functions as training objectives intuitively. Contrastive loss, triplet loss \cite{schroff2015facenet} and margin loss \cite{roth2020revisiting} are among the most fundamental similarity based loss functions. The contrastive loss minimizes objects' distance in the embedding space if their classes are the same, and separates them a fixed margin away otherwise. The triplet loss takes triplets of anchor, positive, and negative objects, and enforces the distance between the anchor and the positive to be smaller than that between the anchor and the negative. Margin loss further extends the triplet loss by introducing a dynamic, learnable boundary $\beta$ between positive and negative pairs. The formation of contrastive loss is as the following. We first have embedding pairs $\mathcal{P}$, which is sampled from a minibatch of size $b$. The pair contains an anchor $\phi_{a}$ from class $y_{a}$ and either a positive $\phi_{p}$ with $y_{a}=y_{p}$ or a negative $\phi_{n}$ from a different class, $y_{a}\neq y_{n}$. The distance function we utilize is the standard Euclidean distance $d_{e}(x, y)=\|x-y\|_{2}$. $\gamma$ is the margin which is usually set to $1$. Then the network $\phi$ is trained to minimize:
\begin{equation}
\mathcal{L}_{\textrm {contrastive}}=\frac{1}{b} \sum_{(i, j) \in \mathcal{P}}^{b}  \mathbb{I}_{y_{i}=y_{j}} d_{e}\left(\phi_{i}, \phi_{j}\right) + \mathbb{I}_{y_{i} \neq y_{j}}\left[\gamma-d_{e}\left(\phi_{i}, \phi_{j}\right)\right]_{+}
\end{equation}
Triplets extend the contrastive formulation by providing a triplet $\mathcal{T}$ sampled from a mini-batch:
\begin{equation}
\mathcal{L}_{\textrm {triplet}}=\frac{1}{b} \sum_{(a, p, n) \in \mathcal{T} \atop y_{a}=y_{p} \neq y_{n}}^{b}\left[d_{e}\left(\phi_{a}, \phi_{p}\right)-d_{e}\left(\phi_{a}, \phi_{n}\right)+\gamma\right]_{+}
\end{equation}
Margin loss transfers the common triplet ranking problem to a relative ordering of pairs: 
\begin{equation}
\mathcal{L}_{\textrm {margin}} = \sum_{(i, j) \in \mathcal{P}}  \gamma +\mathbb{I}_{y_{i}=y_{j}}\left(d_{e}\left(\phi_{i}, \phi_{j}\right)-\beta\right) -\mathbb{I}_{y_{i} \neq y_{j}}\left(d_{e}\left(\phi_{i}, \phi_{j}\right)-\beta\right)
\end{equation}
Given the above formulations, choosing samples becomes the critical step for optimal performance, thus why we introduce reinforcement learning sampling method ASR.
\begin{itemize}
    \item \noindent \textbf{State Space} We use state $s \in S$ to represent current embedding $\phi$. For generalization performance, we use a weighted sum over Recall, NMI on the validation set $\mathcal{O}_{\mathrm{val}}$ to describe $s$.
    \item \noindent \textbf{Action Space and Optimal Policy} We build the action space based on the idea of distance based sampling \cite{wu2017sampling}. \cite{wu2017sampling} proposes to sample triplets from a static uniform distribution over distance between anchor and negative object $d_{an}$: $O_{n}\sim p(O_n|O_a)$. Then we construct a discrete set of parameters to multiply on the probabilities. In different training state $s$, the optimal policy $\pi_{\theta}(a\mid s)$ controls the adjustment to apply.
    \item \noindent \textbf{Reward Function} A reward function is a function that provides a numerical score based on the state of the environment. It maps each state of the environment to a single number, specifying the intrinsic desirability of that state. As we mentioned in Section 3.1, maximizing the reward is the goal in a reinforcement learning algorithm. Formally, we have the \textit{The Reward Hypothesis}: all of what we mean by goals can be thought as the maximization of the expected value of the cumulative sum of a received scalar reward \cite{sutton2018reinforcement}. We compute the reward $r$ for action $a \sim \pi_{\theta}(a \mid s)$ by directly calculating the difference of $\phi(\cdot ; \zeta)$ over $\phi\left(\cdot ; \zeta^{\prime}\right)$ from previous training state.
    \item \noindent \textbf{Cumulative Reward and Policy Objective} In the last part, we model the reward $r$ in an incremental way, however, the goal is to optimize the overall cumulative reward. Thus we introduce the total expected reward $R(\tau)=\sum_{t} r_{t}$. Here $\tau$ is a sequence of training triple for $t = 0,1...,T$. We use $G(s,a)$ to denote the discounted reward, then we can use policy gradient to optimize the policy objective $J(\theta)$ \cite{yu2019convergent} as 
\begin{equation}
\nabla_{\theta} J(\theta)=\mathbb{E}_{\tau \sim \pi_{\theta}(\tau)}\left[\nabla_{\theta} \log \pi_{\theta}(a \mid s) G(s, a)\right]
\end{equation}
\end{itemize}
Computing the gradient $J(\theta)$ is tricky since it depends on both the action space and the final stationary distribution of state space. 
The immediate policy gradient learning algorithm is the REINFORCE algorithm (vanilla policy gradient) \cite{sutton2018reinforcement, agarwal2019reinforcement}. But REINFORCE has two glaring issues: noisy gradients and high variance. Proximal Policy Optimization (PPO) \cite{schulman2017proximal} addresses the above issues by using a surrogate objective to represent the policy objective $J(\theta)$. Thus we use PPO as our premier optimization method but also compare it with REINFORCE in the empirical studies.
\begin{figure*}[!t]
\centering
\includegraphics[width=0.33\linewidth]{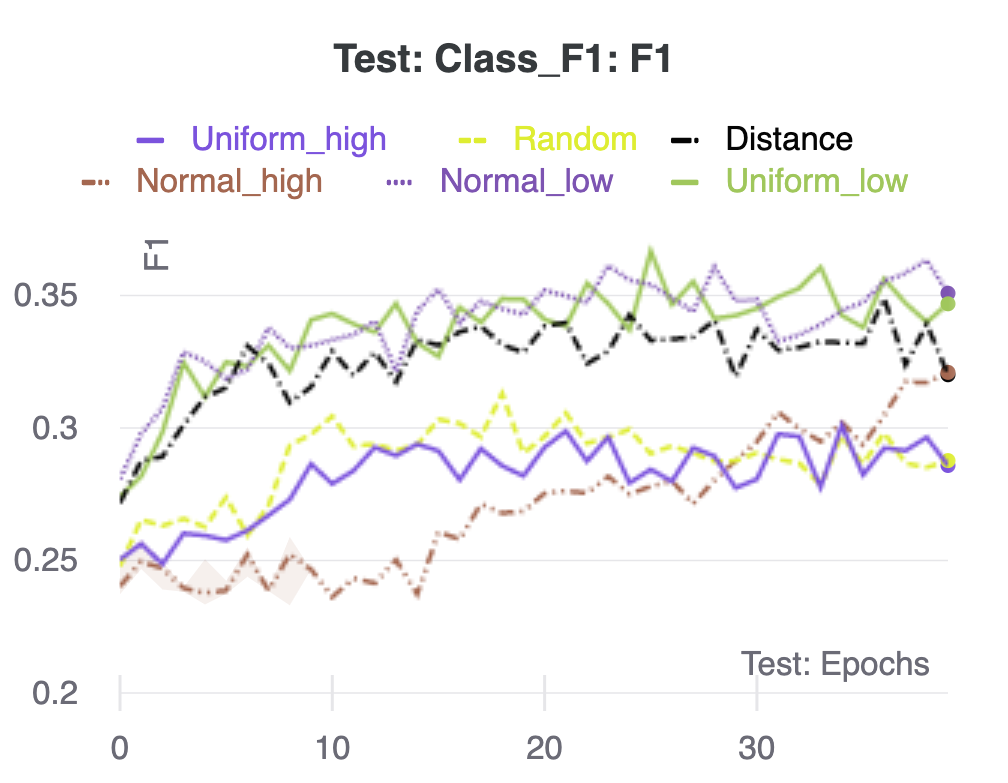}
\includegraphics[width=0.33\linewidth]{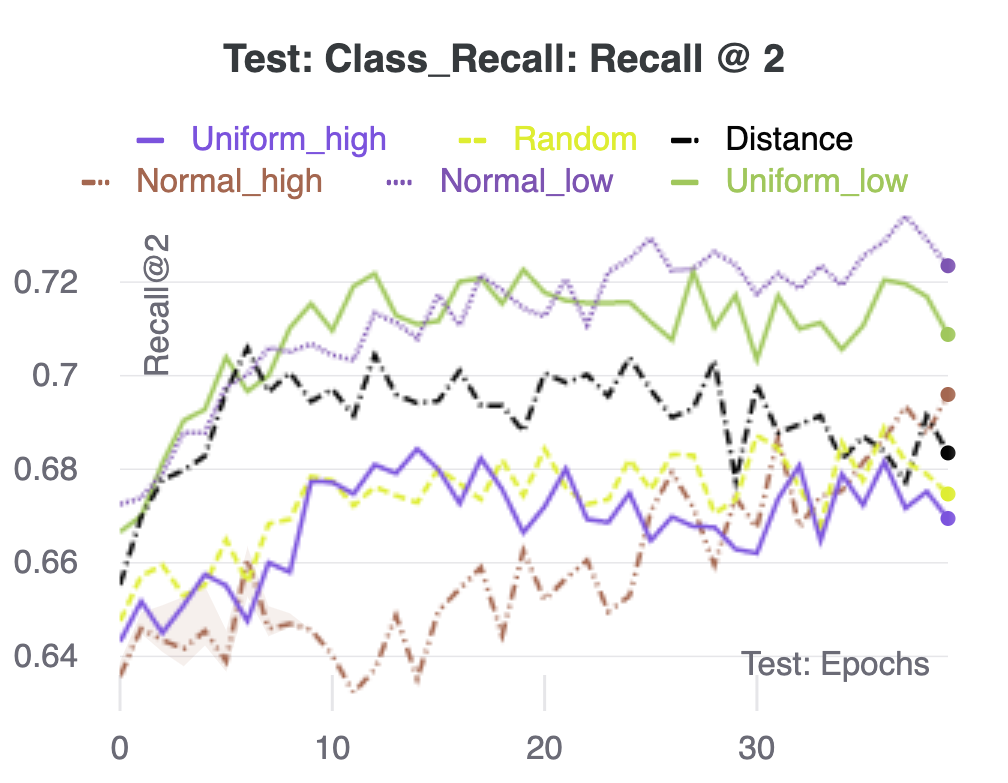}
\includegraphics[width=0.33\linewidth]{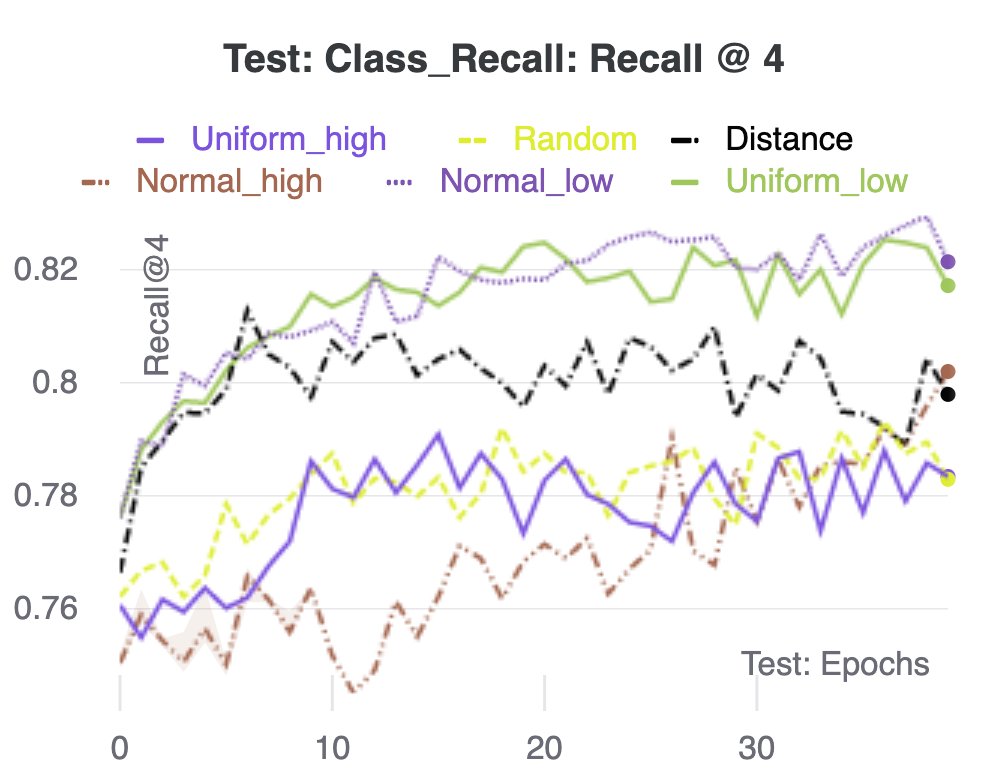}
\includegraphics[width=0.33\linewidth]{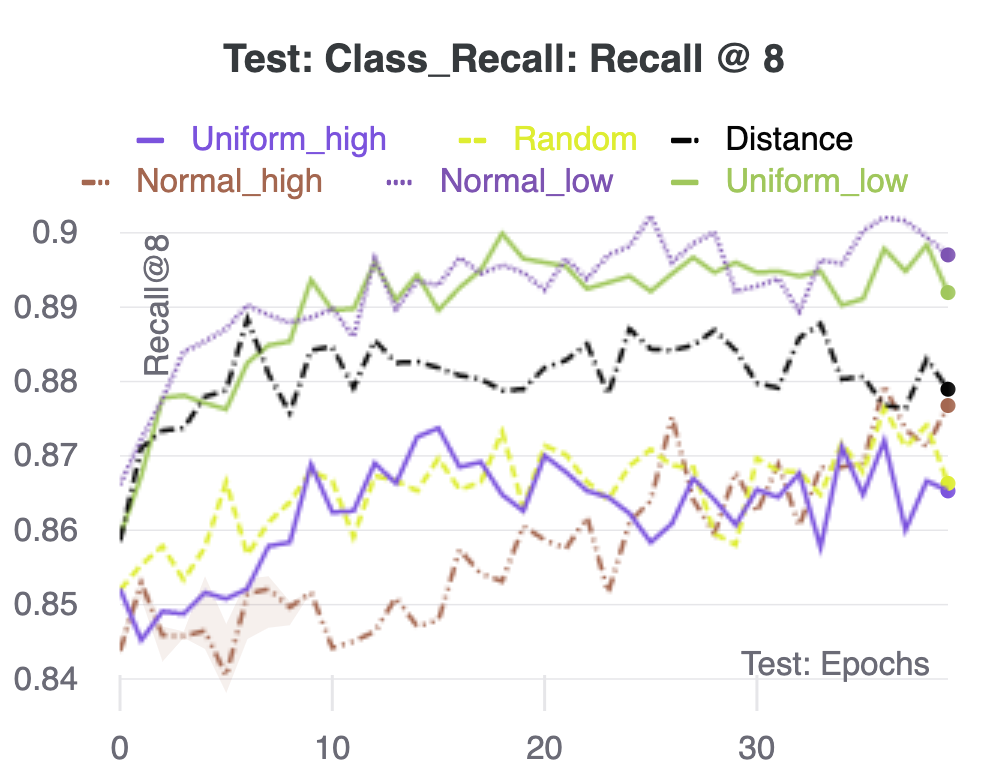}
\includegraphics[width=0.33\linewidth]{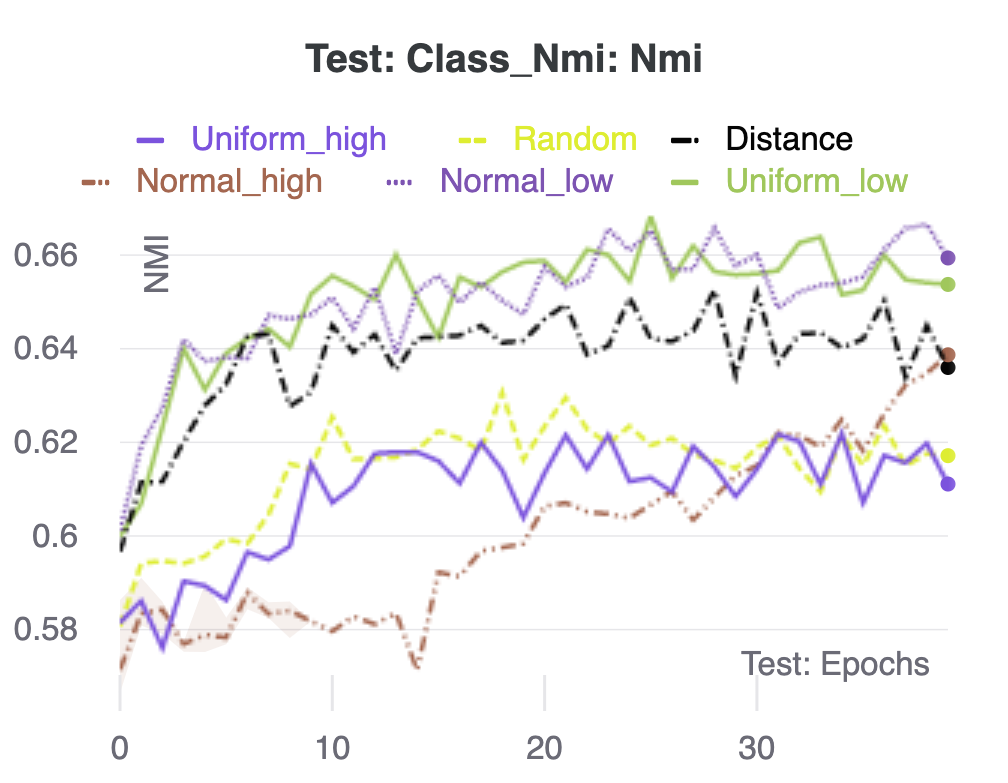}
\includegraphics[width=0.33\linewidth]{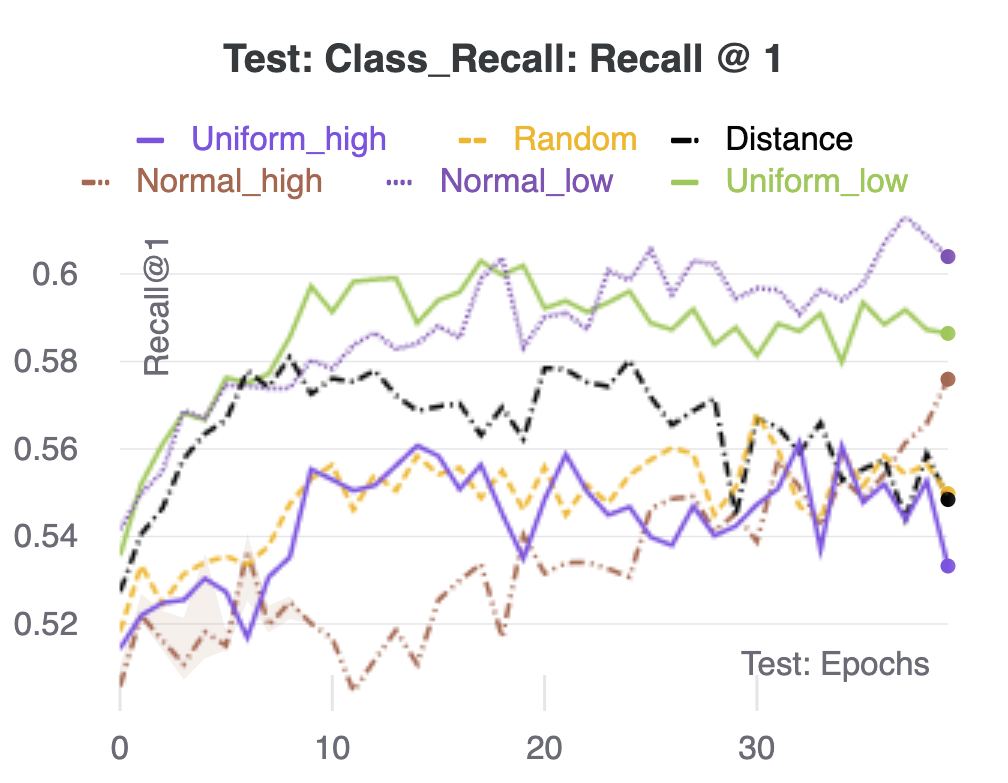}
\caption{Ablation on initial distribution on CUB dataset for six different benchmarks. We observe the ``ASR gravity well'' from the ``Normal\_high'' curves.}
\end{figure*}

The ASR framework is as the following. First, the sampling distribution $p(O_n|O_a)$ is initialized with an initial distribution: random, distance, or Gaussian distribution. Second, we initialize both the adjustment policy $\pi_{\theta}$ and the auxiliary policy $\pi_{\theta_{\mathrm{old}}}$ for estimating the PPO probability ratio. Then training is performed with triplets with random anchor-positive pairs and sampled negatives from the sampling distribution. After selected iterations, reward and state metrics are computed on the embeddings $\phi(\cdot ; \zeta)$ of $\mathcal{O}_{val}$. These values are aggregated in a training reward $r$ and input state $s$. Then $r$ is used to update the current policy $\pi_{\theta}$ and $s$. Then $\pi_{\theta}$ and $s$ are fed into the updated policy to estimate adjustments $a$ for the sampling distribution $p(O_{n}| O_{a})$. After certain iterations $\pi_{\theta_{\mathrm{old}}}$ is updated with current policy $\pi_{\theta}$.

To summarize, for the System 2 approach, we present an adaptive sampling setting for adjustments which directly facilitate maximal representation learning performance. We show how to learn a policy which effectively alters sampling distribution to optimally support learning of the representation models.
\begin{table}[t]
\centering
\setlength{\tabcolsep}{20pt}
\renewcommand{\arraystretch}{1}
\begin{tabular}{c|c}
 \hline
 \textbf{Distribution}&\textbf{parameters}\\
 \hline
 uniform low& $\mu = 0.5, \sigma = 0.2$\\
  \hline
 uniform high& $\mu = 1.5, \sigma = 0.2$\\
  \hline
 distance&limit of distance $=0.5$\\
 \hline
 random& random\\
  \hline
 normal high&$\mu=1.6, \sigma=0.04$\\
  \hline
 normal low& $\mu=0.5, \sigma=0.05$ \\
  \hline
 \end{tabular}
 \vspace{3pt}
 \caption{Initial distribution setup in Section 5.3}
\label{tab:booktabs}
\end{table}
\section{Experiments}
We compare the ASR framework with the most popular heuristic based sampling strategies: semi-hard \cite{schroff2015facenet}, and distance based sampling \cite{wu2017sampling} with both margin and triplet losses on the tasks of retrieval and clustering. For optimization methods, we use both PPO and REINFOCE. As we states earlier, Recall, NMI, and F1 are utilized as evaluation metrics. In this section, we first present implementation details, then the benchmark datasets, experimental results, ablations studies, and in the end, quantitative analysis of our approach to information retrieval and clustering tasks.
\subsection{Implementation Details}
In line with standard practices \cite{wu2017sampling} we resize and crop images to 224 × 224 for training and center crop to the same size for evaluation. Weight decay is set to a constant value of $4 \times 10^{-4}$. All the models are implemented in Pytorch. Experiments are performed on individual Nvidia Titan XP GPU with 12 GB memory. Learning rates are set to $10^{-5}$. Policy $\pi$ is implemented as a two-layer fully-connected network with ReLU-nonlinearity. We choose triplet with $\gamma= 0.2$ and margin $\beta = 0.6$. Each training is run over 40 epochs if not specified. We follow the training protocol with ResNet50 \cite{he2016deep} and a $15\%$ subset of training data is used as validation set.
\subsection{Benchmark Datasets}
We use three common benchmark datasets, including CUB200-2011 (CUB) \cite{WelinderEtal2010}, Cars196 \cite{KrauseStarkDengFei-Fei_3DRR2013} and Stanford Online Products (SOP) \cite{songCVPR16}. Specifically, CUB dataset has 11,788 images, which belong to 200 bird classes. CARS196 dataset has 16,185 images from 196 car classes. SOP dataset contains 120,053 product images within 22,634 classes. For each dataset, the first half of classes are used for training, and the other half are used for testing. And a random subset of $15 \%$ of training sets is used as validation.
\begin{figure*}[t]
\centering
\includegraphics[width=0.33\linewidth]{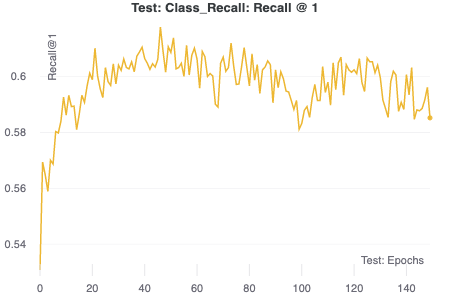}
\includegraphics[width=0.33\linewidth]{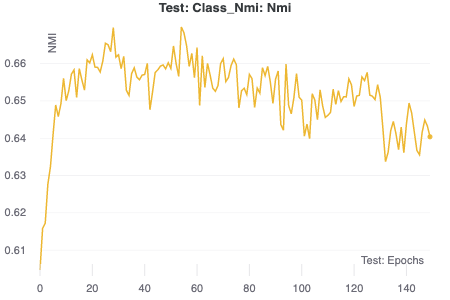}
\includegraphics[width=0.33\linewidth]{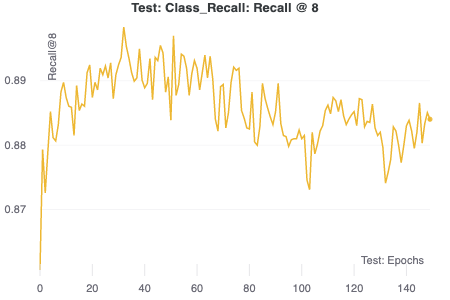}
\includegraphics[width=0.33\linewidth]{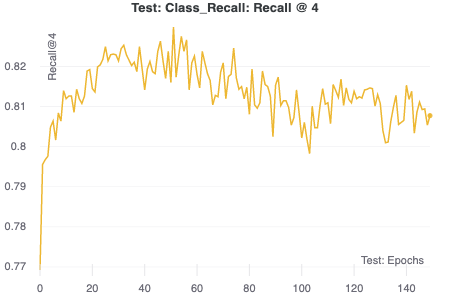}
\includegraphics[width=0.33\linewidth]{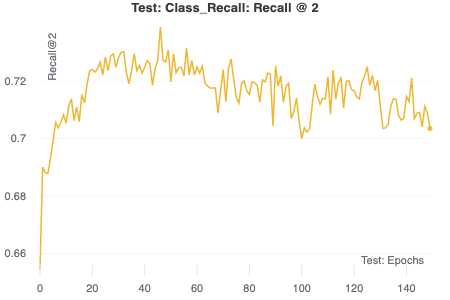}
\includegraphics[width=0.33\linewidth]{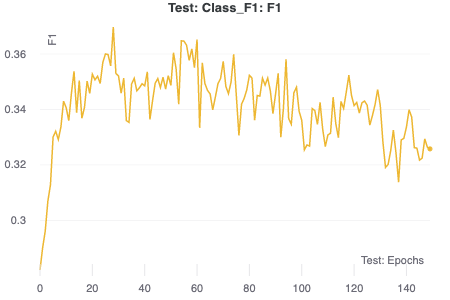}
\caption{Ablation on number of epoch on CUB dataset for Section 5.3}
\end{figure*}
\begin{table*}[h]
\centering
\setlength{\tabcolsep}{3.4mm}
\begin{tabular}{c||c|c|c|c|c|c|c}
 \hline
 \multicolumn{8}{c}{ASR on CUB200-2011~\cite{WelinderEtal2010}} \\
 \hline
 Initial Distribution&R@1&R@2&R@4&R@8&NMI&F1&Loss\\
 \hline
 uniform low&0.5864&0.7129&0.8026&0.8952&0.6539&0.3434&15.084\\
 uniform high&0.543&0.6723&0.7846&0.866&0.6154&0.2912&\textbf{
 1.084}\\
 normal low&\textbf{0.6063}&\textbf{0.7263}&\textbf{0.8255}&\textbf{0.8982}&\textbf{0.6629}&\textbf{0.3571}&14.093\\
 normal high&0.5709&0.6918&0.799&0.874&0.6367&0.319&7.856\\
  random&0.5532&0.6768&0.7862&0.8702&0.6174&0.2864&1.365\\
 distance&0.5537&0.6918&0.8011&0.0881&0.6405&0.3296&19.967\\
  \hline
 \end{tabular}
 \caption{Ablation on initial distribution for CUB200-2011. Training is done over 40 epochs. We can see ``normal high'' is the best initial distribution in most cases.}
\label{tab:booktabs}
\end{table*}
\begin{table}[!t]
\centering
\setlength{\tabcolsep}{2.0pt}
\renewcommand{\arraystretch}{1.}
\begin{tabular}{c||c|c|c|c}
 \hline
 \multicolumn{5}{c}{CUB200-2011 \cite{WelinderEtal2010}} \\
 \hline
 Approach&R@1&R@4&NMI&F1\\
 \hline
  \textbf{ASR(PPO) + Triplet }&\textbf{0.6063}&\textbf{0.8255}&\textbf{0.6629}&\textbf{0.3571}\\
  ASR(PPO) + Margin&0.604&0.8214&0.6594&0.3557\\
  ASR(REINFORCE) + Triplet&0.5037&0.7562&0.6153&0.296\\
 Semihard + Triplet&	0.5947&0.8067&0.6533&0.3412\\
 Distance + Triplet&0.5581&0.7942&0.6262&0.299\\
  \hline
   \multicolumn{5}{c}{CARS196 \cite{KrauseStarkDengFei-Fei_3DRR2013}} \\
 \hline
 &R@1&R@4&NMI&F1\\
 \hline
  \textbf{ASR(PPO) + Triplet}&\textbf{0.7150}&\textbf{0.8930}&0.5993&0.2925\\
    ASR(PPO) + Margin&0.6179&0.839&0.5552&0.2627\\
  ASR(REINFORCE) + Triplet&0.0.7145&0.8901&\textbf{0.602}&\textbf{0.3104}\\
 Semihard + Triplet&0.6633&0.8491&0.5836&0.2665\\
 Distance + Triplet&0.6337&0.8393&0.5508&0.2216\\
  \hline
 \multicolumn{5}{c}{SOP \cite{songCVPR16}} \\
 \hline
 &R@10&R@100&NMI&F1\\
 \hline
  \textbf{ASR(PPO) + Triplet}&\textbf{0.9447}&0.8737&0.8914&\textbf{0.3258}\\
 ASR(PPO) + Margin&0.8619&0.9387&0.8911&0.3247\\
 ASR(REINFORCE) + Triplet&0.85&0.9296&0.8904&0.3217\\
 Semihard + Triplet&0.8703&\textbf{0.9444}&\textbf{0.8916}&0.3265\\
 Distance + Triplet&0.8337&0.926&0.8813&0.2806\\
  \hline
 \end{tabular}
 \vspace{3pt}
 \caption{Comparison of sampling strategies and loss functions. Training is done over 40 epochs. Three datasets are used for comprehensive empirical evaluation. Four benchmarks are used to evaluate models' performance. We can see in most cases ASR approach achieves the best performance.}
\label{tab:booktabs}
\end{table}
\subsection{Results and Analysis}
Empirical results in Table 3 demonstrate that reinforcement learning approach to the adaptive sampling effectively supports better learning of the representations and achieves superior performance on information retrieval and clustering tasks. We can see ASR approach outperforms the heuristic-based sampling approach in all three benchmark datasets.

\cite{mei2020escaping} suggests that optimizing with respect to the softmax exhibits sensitivity to parameter initialization. So we explore different distribution initialization and study the impact of the number of epoch in the following subsections on the CUB dataset. We utilize a variety of distributions named ``uniform'', ``distance'', ``random'', and ``normal'' to study policy initiation's impact on ASR framework's performance. Table 1 contains detailed distribution parameters setting.

\subsubsection{Policy Initialization: ASR Gravity Well}
\begin{table}[t]
\centering
\setlength{\tabcolsep}{1.1mm}
\renewcommand{\arraystretch}{1.1}
\begin{tabular}{c||c|c|c|c|c|c}
 \hline
 \textbf{Measurement}&F1&R@1&R@2&R@4&R@8&NMI\\
 \hline
 \textbf{Epoch}&28&46&46&51&32&28\\
  \hline
 \end{tabular}
  \vspace{3pt}
 \caption{Statistics of the optimal number of epoch on CUB dataset. Training is done over 150 epochs.}
\label{tab:booktabs}
\vspace{-0.6cm}
\end{table}
We notice a captivating phenomenon which we name as ``ASR gravity well'' in the experiments. ``Softmax gravity well''(SGW) \cite{mei2020escaping} describes such a phenomenon: gradient ascent trajectories are drawn toward suboptimal corners of the probability simplex and subsequently slowed in their progress toward the optimal vertex. These facts are observed through empirical observations in one-state MDPs, Four-room environment, and MNIST dataset in \cite{mei2020escaping}, revealing that the behavior of Softmax policy gradient depends strongly on initialization. We observe a similar phenomenon in our experiments with ``relative high'' normal distribution. When we closely investigate the ``Normal\_high'' curves in Figure 2, we find they all fall into a global minimum at around epoch 15. The ``reverse peaks'' are very significant within five (Recall@1, Recall@2, Recall@4, Recall@8, and NMI) out of six benchmarks. That means the phenomenon arises for both information retrieval and clustering applications. Since ASR adopts softmax policy gradient, we name it as ``ASR gravity well''. ``Softmax gravity well'' \cite{mei2020escaping} captures that poorly initialized softmax policy gradient algorithm can get stuck at suboptimal policy for single-state MDP. We confirm and extend this intriguing observation to practical deep reinforcement learning's application in the sampling problem of representation learning.

For the normal distribution, when the $\mu$ is high, it tends to change dramatically, so it has a high tendency to trap into the ``ASR gravity well''. We also notice ``normal low'' and ``uniform low'' are the two distributions that perform the best from Table 2. It's intuitive since then we have smaller $\sigma$, a normal distribution skew towards uniform distribution. And actions with larger true rewards will have larger gradients in normal low/uniform distribution therefore the attraction towards suboptimal corners will not happen. It would be interesting to see how the phenomenon evolves in other scenarios like attention models and exponential exploration.
\subsubsection{Number of Epoch's Impact}
We study the optimal number of epochs to train with CUB dataset by ASR framework and triplet loss. The curves of experimental results are shown in Figure 3 and the numerical performances are summarized in Table 4. We find our method usually achieves the best performance between 30 epochs and 50 epochs. This confirms our practice that different comparison methods all run 40 epochs to reach (approximate) optimum in experiments. Increasing number of epochs can cause the over-fitting problem and deteriorate the performance. 
\section{Social Impact and Discussion}
This research pursues  methodology innovation, so our social implications are relatively high level and have no direct negative impact. While machine learning technologies have consumed significant energy footprint, our research can increase energy consumption.

Reinforcement learning can help guiding sampling process, however it increases the total number of hyper-parameters required to be tuned. This can be a drawback in comparison to simpler methods like (contextual) bandit and active learning. So ASR framework is the best suitable in machine learning settings where the perspective of long term return instead of the immediate reward (active learning) is critical. We are glad to validate empirically that the representation learning sampling problem is the case.

\section{Conclusion}
We propose the ASR framework first applying reinforcement learning to address adaptive sampling problem in representation learning. The ASR framework represents System 2 Thinking in cognitive science theory and heuristic-based baseline methods represent System 1 thinking. Experiments in representation learning show that ASR with triplet loss has the best performance for most metrics on two applications and three datasets. 

The competitive  performance across different datasets validate our method's generalizability and robustness. Additionally, our work leaves several intriguing questions: natural policy gradient and log-barrier regularization offer nice theoretical convergence properties, how do they compare to proximal policy optimization in the sampling framework \cite{cai2020provably}? How much ``escape time'' around each suboptimal corner does the algorithm spend to be out of the ``gravity well''? Other interesting directions are extending the sampling framework to other areas like few-shot Meta-Learning, and zero-shot Domain Adaptation.


\bibliographystyle{unsrt}  
\bibliography{references}
\end{document}